%% file: example_paper.tex
\documentclass[nohyperref]{article}

\usepackage{microtype}
\usepackage{graphicx}
\usepackage{subfigure}
\usepackage{booktabs} 

\usepackage{hyperref}

\usepackage[accepted]{icml2022}

\usepackage{amsmath}
\usepackage{amssymb}
\usepackage{mathtools}
\usepackage{amsthm}
\usepackage{multirow}
\usepackage{caption}
\usepackage{subfigure}
\usepackage{graphicx}
\usepackage{adjustbox}
\usepackage{enumitem}

\usepackage[capitalize,noabbrev]{cleveref}

\theoremstyle{plain}
\newtheorem{theorem}{Theorem}[section]

\theoremstyle{definition}
\newtheorem{definition}[theorem]{Definition}
\newtheorem{assumption}[theorem]{Assumption}
\theoremstyle{remark}

\newcommand{\norm}[1]{\left\lVert#1\right\rVert}
\newcommand{\abs}[1]{\left|#1\right|}

\usepackage[textsize=tiny]{todonotes}

\newlist{myitems}{enumerate}{3}
\setlist[myitems, 1]
{label=\arabic{myitemsi}.,leftmargin=15pt,labelwidth=10pt,labelsep=5pt,
topsep=0pt,parsep=0pt,partopsep=0pt,noitemsep
}

\icmltitlerunning{Stabilizing GANs' Training with Brownian Motion Controller}

\begin{document}

\twocolumn[
\icmltitle{Stabilizing GANs' Training with Brownian Motion Controller}




\begin{icmlauthorlist}
\icmlauthor{Tianjiao Luo}{thu}
\icmlauthor{Ziyu Zhu}{thu}
\icmlauthor{Jianfei Chen}{thu}
\icmlauthor{Jun Zhu}{thu,pazhou}
\end{icmlauthorlist}

\icmlaffiliation{thu}{Dept. of Comp. Sci. \& Tech., Institute for AI, BNRist Center, Tsinghua-Bosch Joint ML Center, THBI Lab, Tsinghua University}
\icmlaffiliation{pazhou}{Pazhou Lab (Huangpu), Guangzhou, China}


\icmlcorrespondingauthor{Jianfei Chen}{jianfeic@tsinghua.edu.cn}

\icmlkeywords{Machine Learning, ICML}

\vskip 0.3in
]



\printAffiliationsAndNotice{} 

\begin{abstract}
The training process of generative adversarial networks (GANs) is unstable and does not converge globally. In this paper, we examine the stability of GANs from the perspective of control theory and propose a universal higher-order noise-based controller called Brownian Motion Controller (BMC). Starting with the prototypical case of Dirac-GANs, we design a BMC to retrieve precisely the same but reachable optimal equilibrium. We theoretically prove that the training process of DiracGANs-BMC is \textit{globally exponential stable} and derive bounds on the rate of convergence. Then we extend our BMC to normal GANs and provide implementation instructions on GANs-BMC. Our experiments show that our GANs-BMC effectively stabilizes GANs' training under StyleGANv2-ada frameworks with a faster rate of convergence, a smaller range of oscillation, and better performance in terms of FID score.
\end{abstract}

\input{data/1-intro}

\input{data/2-gan}
\input{data/3-preliminary}
\input{data/4-dirac-gan}

\input{data/5-normal-gan}

\input{data/6-experiments}

\section{Conclusion and Discussion}
In this paper, we revisit GANs' instability problem from the perspective of control theory. Our work novelly incorporates noise-based controller of on the training dynamics of GANs and modifies its objective function accordingly to stabilize GANs. We innovatively design a Brownian Motion Control (BMC) to achieve globally exponential stability. Notably, our BMC is compatible with all GANs variations. In our paper, theoretical analysis has been done under the Dirac-GANs setting, and we are able to stabilize both the generator and discriminator simultaneously. Experimental results demonstrate that our BMC accelerates the convergence of GANs and performs better in terms of FID than StyleGANv2-ada on CIFAR-10, LSUN-Bedroom, LSUN-Cat, and FFHQ. 

\textbf{Possible Future Directions} While in practice, the training process of GANs is discrete, we model it as a continuous dynamic system in our work. As a result, our method performs better on training with smaller time intervals per iteration. Additionally, although many advanced control methods, including our BMC, are available to stabilize the dynamic system, few of them have been applied in deep generative models. As a result, future work can be done on modeling training as a discrete process and exploring more control methods to improve its stability. 

\textbf{Social Impacts and Potential Harmful Consequences} Like other generative models, our GAN-BMC has ethical problems related to deep fake. The generated photos and videos are indistinguishable from the real ones and may be used illegally to spread misleading information for harmful uses. Secondly, our GAN-BMC may be biased toward concern identity categories, such as race and gender. Product farming is another ethical concern GAN-BMC has, as generated images join art contests and bring fairness issues for artists.

\section*{Acknowledgements}
This work was supported by the National Key Research and Development Program of China
(2021ZD0110502); 
NSF of China Projects (Nos. 62061136001, 61620106010, 62076145, U19B2034, U1811461, U19A2081, 6197222, 62106120, 62076145); a grant from Tsinghua Institute for Guo Qiang; the High Performance Computing Center, Tsinghua University. J.Z was also supported by the New Cornerstone Science Foundation through the XPLORER PRIZE. We thank Gabriele Oliaro from Carnegie Mellon University for his assistance on an earlier draft.

\bibliography{example_paper}
\bibliographystyle{icml2022}

\newpage
\appendix
\onecolumn

\section*{Appendix A: Proof of Theorem 5.2}
 Under Assumption \ref{ass2}, for any initial value $X(0)=\xi\in  \mathbb{R}^{2}$, if $\varrho_{2}\neq 0$ and $\beta>1$, then there a.s. exists a unique global solution $X(t)$ to system (\ref{2.2}) on $t\in[0, \infty)$.
\begin{proof}
Under Assumption \ref{ass2}, then, we can calculate that
 \begin{align*}\label{2.1}
   &X^{\mathrm{T}}(t)f(X(t))\\\nonumber
   =&\phi(t)h^{'}_1(\phi(t)c)c+\phi(t)h^{'}_2(\phi(t)(\tilde{\theta}(t)+c))\tilde{\theta}(t)+\\\nonumber
   &\phi(t)h^{'}_2(\phi(t)(\tilde{\theta}(t)+c))c +\\\nonumber
   &\tilde{\theta}(t)h^{'}_3(\phi(t)(\tilde{\theta}(t)+c))\phi(t)\\\nonumber
   \leq&[(1+\frac{1}{2}\alpha_{1}^{2})c^{2}+2c+\frac{1}{2}]|X|^{2}+(\alpha_{2}^{2}+\frac{1}{2}\alpha_{3}^{2})|X|^{4}.
 \end{align*}
For any bounded initial value $X(0)\in \mathbb{R}^{n}$, there exists a unique maximal local strong solution $X(t)$ of system (\ref{2.2}) on $t\in[0, \tau_{e})$, where $\tau_{e}$ is the explosion time. To show that the solution is actually global, we only need to prove that $\tau_{e}=\infty$ a.s. Let $k_{0}$ be a sufficiently large positive number such that $|X(0)|<k_{0}$. For each integer $k\geq k_{0}$, define the stopping time
\begin{equation*}
\begin{split}
\begin{aligned}
\tau_{k}=\inf\{t\in[0, \tau_{e}):|X(t)|\geq k\}
\end{aligned}
\end{split}
\end{equation*}
with the traditional setting $\inf\emptyset=\infty$, where $\emptyset$ denotes the empty set. Clearly, $\tau_{k}$ is increasing as $k\rightarrow\infty$ and $\tau_{k}\rightarrow\tau_{\infty}\leq\tau_{e}$ a.s. If we can show that $\tau_{\infty}=\infty$, then $\tau_{e}=\infty$ a.s., which implies the desired result. This is also equivalent to prove that, for any $t>0$, $\mathbb{P}(\tau_{k}\leq t)\rightarrow 0$ as $k\rightarrow\infty$. To prove this statement, for any $p\in(0,1)$, define a $C^{2}$-function
\begin{equation*}
\begin{split}
\begin{aligned}
V(x)=|X(t)|^{p}.
\end{aligned}
\end{split}
\end{equation*}
One can obtain that $X(t)\neq 0$ for all $0\leq t\leq \tau_{e}$ a.s. Thus, one can apply the It$\mathrm{\hat{o}}$ formula to show that for any $t\in[0, \tau_{e})$,
\begin{equation*}
\begin{split}
\begin{aligned}
\mathrm{d}V(X(t))=&LV(X(t))\mathrm{d}t+p\varrho_{1}|X(t)|^{p}\mathrm{d}B_{1}(t)\\
&+p\varrho_{2}|X(t)|^{\beta+p}\mathrm{d}B_{2}(t),
\end{aligned}
\end{split}
\end{equation*}
where $LV$ is defined as
\begin{equation*}
\begin{split}
\begin{aligned}
LV(X)=&p|X|^{p-2}X^{\mathrm{T}}f(X(t))+{\frac{p(p-1)\varrho_{1}^{2}}{2}}|X|^{p}\\
&+\frac{p(p-1)\varrho_{2}^{2}}{2}|X|^{2\beta+p}
\end{aligned}
\end{split}
\end{equation*}
By Assumption \ref{ass2}, we therefore have
\begin{equation*}
\begin{aligned}
LV(X)\leq&\frac{p(p-1)\varrho_{2}^{2}}{2}|X|^{2\beta+p}+((1+\frac{1}{2}\alpha_{1}^{2})c^{2}+2c+\frac{1}{2}) \\p|X|^{\alpha+p} &
+p\Bigg(\frac{(p-1)\varrho_{1}^{2}}{2}+(\alpha_{2}^{2}+\frac{1}{2}\alpha_{3}^{2})\Bigg)|X|^{p}.
\end{aligned}
\end{equation*}

Noting that $p\in(0,1)$ and $\beta>1$ and $\varrho_{2}\neq 0$, by the boundedness of polynomial functions, there exists a positive constant $\bar{H}$ such that $LV(x)\leq \bar{H}$. We therefore have
\begin{equation*}
\begin{split}
\begin{aligned}
\mathbb{E}V(X(t\wedge\tau_{k}))&\leq\mathbb{E}|\xi|^{p}+\mathbb{E}\int_{0}^{t\wedge\tau_{k}}LV(X(s))\mathrm{d}s\\
&\leq\mathbb{E}|\xi|^{p}+\bar{H}_{t}\\
&=:\bar{H}_{t},
\end{aligned}
\end{split}
\end{equation*}
where $\bar{H}_{t}$ is independent of $k$. By the definition of $\tau_{k}$, $|X(\tau_{k})|=k$, so
\begin{equation*}
\begin{split}
\begin{aligned}
\mathbb{P}(\tau_{k}\leq t)k^{p}&\leq\mathbb{P}(\tau_{k}\leq t)V(X(\tau_{k}))]\\
&\leq\mathbb{E}[l_{\{\tau_{k}\leq t\}}V(X(t\wedge\tau_{k}))]\\
&\leq\mathbb{E}V(X(t\wedge\tau_{k}))\\
&\leq\bar{H}_{t},
\end{aligned}
\end{split}
\end{equation*}
which implies that
\begin{equation*}
\begin{split}
\begin{aligned}
\limsup_{k\rightarrow\infty}\mathbb{P}(\tau_{k}\leq t)\leq \lim_{k\rightarrow\infty}\frac{\bar{H}_{t}}{k^{p}}=0,
\end{aligned}
\end{split}
\end{equation*}
as required.
\end{proof}
\newpage

\section*{Appendix B: Proof of Theorem 5.3}
 Let Assumption \ref{ass2} hold. Assume that $\varrho_{2}\neq 0$ and $\beta>1$. If
\begin{equation*}
\begin{split}
\begin{aligned}
\frac{\varrho_{1}^{2}}{2}-\varphi>0,
\end{aligned}
\end{split}
\end{equation*}
where
\begin{equation}
\begin{split}
\begin{aligned}
\varphi=\max_{x\geq 0}\Bigg\{-\frac{\varrho_{2}^{2}}{2}x^{2\beta}+(\alpha_{2}^{2}+\frac{1}{2}\alpha_{3}^{2})x^{2}+[(1+\frac{1}{2}\alpha_{1}^{2})c^{2}+2c+\frac{1}{2}]\Bigg\},
\end{aligned}
\end{split}
\end{equation}
then for any $X(0)=\xi$ with sufficiently small constant $\epsilon\in(0, \varrho_{1}^{2}/2-\varphi)$, the global solution $X(t)$ of system (\ref{2.2}) has the property that
\begin{equation*}
\begin{split}
\begin{aligned}
\limsup_{t\rightarrow\infty}\frac{\log|X(t)|}{t}\leq-\bigg(\frac{\varrho_{1}^{2}}{2}-\varphi\bigg)+\epsilon,~~~~a.s.
\end{aligned}
\end{split}
\end{equation*}
that is, the solution of system (\ref{2.2}) is a.s. exponentially stable.

\begin{proof}
Applying It$\mathrm{\hat{o}}$ formula to $\log|X(t)|$ yields
\begin{equation*}
\begin{split}
\begin{aligned}
\log|X(t)|=&\log|X(0)|+\int_{0}^{t}\Bigg[|X(t)|^{-2}X^{\mathrm{T}}(s)f(X(s))\\
&-\frac{\varrho_{2}^{2}}{2}|X(s)|^{2\beta}-\frac{\varrho_{1}^{2}}{2}\Bigg]\mathrm{d}s+\int_{0}^{t}\varrho_{1}\mathrm{d}B_{1}(t)\\
&+\varrho_{2}\int_{0}^{t}|X(s)|^{\beta}\mathrm{d}B_{2}(s).
\end{aligned}
\end{split}
\end{equation*}
Letting $M(t)=\varrho_{2}\int_{0}^{t}|X(s)|^{\beta}\mathrm{d}B_{2}(s)$, clearly $M(t)$ is a continuous local martingale with the quadratic variation
\begin{equation*}
\begin{split}
\begin{aligned}
<M(t),M(t)>=\varrho_{2}^{2}\int_{0}^{t}|X(s)|^{2\beta}\mathrm{d}s.
\end{aligned}
\end{split}
\end{equation*}
For any $\varepsilon\in(0,1)$, choose $\theta>0$ such that $\theta\varepsilon>1$. Then for each integer $m>0$, the exponential martingale inequality gives
$$\mathbb{P}\Bigg\{\sup_{1\leq t\leq m}\Bigg[M(t)-\frac{\varepsilon\varrho_{2}^{2}}{2}\int_{0}^{t}|X(s)|^{2\beta}\mathrm{d}s\Bigg]\geq\theta\varepsilon\log m\Bigg\}\leq\frac{1}{m^{\theta\varepsilon}}.$$
Since $\sum_{m=1}^{\infty}m^{-\theta\varepsilon}<\infty$, by the well-known Borel-Cantelli lemma, there exists an $\bar{\Omega}_{0}\subseteq\Omega$ with $\mathbb{P}(\bar{\Omega}_{0})=1$ such that for any $\omega\in\bar{\Omega}_{0}$, there exists an integer $\bar{m}(\omega)$, when $m>\bar{m}(\omega)$, and $m-1\leq t\leq m$,
\begin{equation*}
\begin{split}
\begin{aligned}
M(t)\leq\frac{\varepsilon\varrho_{2}^{2}}{2}\int_{0}^{t}|X(s)|^{2\beta}\mathrm{d}s+\theta\varepsilon\log(t+1).
\end{aligned}
\end{split}
\end{equation*}
This, together with Assumption \ref{ass2}, yields
\begin{equation*}
\begin{split}
\begin{aligned}
\log|X(t)|\leq&\log|\xi|+\int_{0}^{t}\Bigg[-\frac{\varrho_{2}^{2}(1-\varepsilon)}{2}|X(s)|^{2\beta}\\
&+(1+(\frac{1}{2}\alpha_{1}^{2})c^{2}+2c+\frac{1}{2})|X(s)|^{\alpha}\\
&+(\alpha_{2}^{2}+\frac{1}{2}\alpha_{3}^{2})-\frac{\varrho_{1}^{2}}{2}\Bigg]\mathrm{d}s\\
&+\int_{0}^{t}\varrho_{1}\mathrm{d}B_{1}(t)+\theta\varepsilon\log(t+1).
\end{aligned}
\end{split}
\end{equation*}
Letting $\epsilon$ be sufficiently small, by the definition of $\varphi$ in (\ref{fai}), for sufficiently small $\epsilon\in(0,\varrho_{1}^{2}/2-\varphi)$, we have
\begin{equation*}
\begin{split}
\begin{aligned}
\log|X(t)|\leq&\log|\xi|-\Bigg[(\frac{\varrho_{1}^{2}}{2}-\varphi)-\epsilon\Bigg]t+\int_{0}^{t}\varrho_{1}\mathrm{d}B_{1}(t)\\
&+\theta\varepsilon\log(t+1).
\end{aligned}
\end{split}
\end{equation*}
Applying the strong law of large number, we therefore have
\begin{equation*}
\begin{split}
\begin{aligned}
\limsup_{t\rightarrow\infty}\frac{\log|X(t)|}{t}\leq-(\frac{\varrho_{1}^{2}}{2}-\varphi)+\epsilon~~~~a.s.
\end{aligned}
\end{split}
\end{equation*}
\end{proof}
\newpage

\section*{Appendix C: Experimental Setups}

\begin{table}[htpb!]
  \caption{ Detailed Experiment Setups of StyleGANv2-ada and its BMC trails}
    \centering
    \begin{tabular}{c|c|c|c|c}
    \toprule
    Dataset  &  Batch Size & Learning Rate & Optimizer & GPUs\\
    \hline
      CIFAR-10	   & 64 & 0.0025	&Adam	&2\\
      LSUN-Cat (256x256) &	64	&0.0025	&Adam	&4 \\
      LSUN-Bedroom (256x256)	&64	&0.0025	&Adam	&4 \\
      FFHQ (1024x1024)&	32	&0.0002	&Adam	&4 \\
    \hline
    \end{tabular}
    \label{table:style_setup}
\end{table}

\begin{table}[htpb!]
    \centering
    \caption{Detailed Experiment Setups of ProjectedGAN and its BMC trails}
    \begin{tabular}{c|c}
    \toprule
       Dataset  & CIFAR-10 \\
       \hline
       Batch Size  & 64\\
       \hline
       Learning Rate & 0.002\\
       \hline
       Optimizer & Adam \\
       \hline
       Projected & EfficientNet\\
    \hline
    \end{tabular}
    \label{tab:porjected_setup}
\end{table}
\newpage

\section*{Appendix D: Results on ProjectedGAN and ImageNet}

\begin{table}[htpb!]
    \centering
    \caption{FID scores on CIFAR-10 for ProjectedGAN with respect to the number of real images seen}
    \begin{tabular}{c|c|c|c|c|c}
    \toprule
        & 200k&	400k&	600k&	800k&	1M  \\
    \hline
 Projected-GAN&	12.77&	9.40&	8.17&	7.34&	7.01\\
 Projected-GAN-BMC ($\varrho_1
=1$, 
$\varrho_2=0.1$)&	11.20&	9.11&	7.28&	6.67&	6.28\\
Projected-GAN-BMC ($\varrho_1
=0.1$, 
$\varrho_2=0.01$)&	12.57&	9.31&	8.17&	7.31&	6.74\\
Projected-GAN-BMC ($\varrho_1
=0.01$, 
$\varrho=0.001$)&	12.17&	8.92&	7.93&	6.42&	6.39\\
\hline
    \end{tabular}
    
    \label{tab:my_label}
\end{table}

\begin{table}[htpb!]
    \centering
    \caption{ FID scores on ImageNet 32 $\times$
 32 with respect to the number of real images seen}
    \begin{tabular}{c|c|c|c|c|c|c}
    \toprule
         &  1M & 2M & 3M& 4M & 5M & 6M\\
         \hline
  StyleGANv2-ada	&28.67	&18.93	&15.94 &14.17&	13.46&	12.45\\
StyleGANv2-ada-BMC ($\varrho_1 =1$ , $\varrho_2=0.1$)&	21.41&	16.10&	13.76&	12.25&	11.48&	10.93\\
StyleGANv2-ada-BMC ($\varrho_1
 = 0.1$ , 
$\varrho_2=0.01$)&	25.34&	17.31&	14.84&	12.89&	12.10&	11.79\\
StyleGANv2-ada-BMC ($\varrho_1
 = 0.01$ , 
$\varrho_2=0.001$)&26.85&	17.89&	15.46&	13.50&	12.88&	12.31\\
\hline
\end{tabular}
 \label{tab:img_net}
\end{table}

\end{document}

%% file: data/1-intro.tex
\section{Introduction}
\begin{table}[h]
 \caption{Summarization of the converging behaviors under Dirac-GANs' setting on various GANs from existing theoretical stability analysis \citep{mescheder2018training, Fedus2018diverge,Farnia2020nash}. Notice that $\checkmark^{*}$ indicates observations from our experiments.}
    \begin{adjustbox}{width={82mm},keepaspectratio}%
    \centering
    \begin{tabular}{ccccc}
    \toprule
      & Unstable   &  {\begin{tabular}[c]{@{}c@{}}Local\\Stability \end{tabular}}  & {\begin{tabular}[c]{@{}c@{}}Global\\Stability \end{tabular}} \\
    \midrule
       
       {\begin{tabular}[c]{@{}c@{}}Original\\ \citep{goodfellow2014gan} \end{tabular}}  &\checkmark & & \\
       {\begin{tabular}[c]{@{}c@{}}WGAN\\ \citep{Arjovsky2017trainingmethod} \end{tabular}}  &\checkmark & &\\
       {\begin{tabular}[c]{@{}c@{}}WGAN-GP\\ \citep{Gulrajani12017improvewgan} \end{tabular}} &\checkmark & &\\
       {\begin{tabular}[c]{@{}c@{}}DRAGAN\\ \citep{Kodaliconvergestability} \end{tabular}}  &\checkmark & & \\
       {\begin{tabular}[c]{@{}c@{}}Instance Noise\\ \citep{sonderby2016amortised} \end{tabular}}  & &\checkmark & \\
       {\begin{tabular}[c]{@{}c@{}}CLC\\ \citep{xu2019understanding} \end{tabular}}  & &\checkmark & \\
       {\begin{tabular}[c]{@{}c@{}}TTUR\\ \citep{Heuse2017localnash} \end{tabular}}  & &\checkmark & \\
       {\begin{tabular}[c]{@{}c@{}}StyleGAN\\ \citep{karras2019style,karras2020training,karras2021alias} \end{tabular}}  & $\checkmark^{*}$& & \\
       \midrule
     \textbf{Our Method} & & \checkmark & \checkmark \\
     \bottomrule
    \end{tabular}
    \end{adjustbox}
   
    \label{tab:my_label}   
\end{table}
Generative Adversarial Networks (GANs) \citep{goodfellow2014gan} are popular deep generative models. Given a multi-dimensional input dataset with an unknown distribution $\displaystyle p(x)$, GANs can obtain an estimated $\displaystyle p_{G}(x)$ and produce new samples that are almost indistinguishable from the training samples. 
GANs have many application on image generation~\citep{brock2018large}, representation learning~\citep{radford2015unsupervised,salimans2016improved,mathieu2016disentangling}, and image to image translation~\citep{zhu2017unpaired}.

GANs consist of two neural networks: a \textit{generator} and a \textit{discriminator}. 
The generator creates new samples that resemble those from the input dataset as closely as possible. The discriminator, on the other hand, aims to distinguish the (counterfeit) samples produced by the generator from the members of the input dataset. The two networks can be modeled as a minimax problem; they compete against one another while striving to reach a \textit{Nash-equilibrium}, an optimal solution where the generator can produce fake samples that are, from the point of view of the discriminator, in all respects indistinguishable from real ones.

Unfortunately, the process of training GANs often suffers from instabilities, preventing them from reaching the Nash equilibrium. 
There are many attempts for stabilizing GANs, including alternative architectures~\cite{miyato2018spectral,brock2018large,karras2019style}, regularizations~\cite{Kodaliconvergestability,karras2020analyzing}, alternative training algorithms~\cite{Heuse2017localnash,karras2020analyzing}, and alternative objective functions~\cite{Gulrajani12017improvewgan,nowozin2016f,li2017mmd}. However, as shown by \citet{mescheder2018training,Farnia2020nash}, though being empirically effective, many of these methods still suffer from instability even for very simple GANs' architectures. 

One promising approach for analyzing and stabilizing GANs' training is through the perspective of dynamical systems~\cite{Arjovsky2017trainingmethod,Kodaliconvergestability,xu2019understanding}. Rather than the original minimax problem, they directly analyze the continuous-time training dynamics, which is a system of ordinary differential equations (ODEs) known as gradient flows. In this formulation, the behavior of GAN training closely relates to the concept of \emph{asymptotic stability}, which depicts whether the training trajectory can eventually converge in the vicinity of an equilibrium of the gradient flow. 
Moreover, the stability of GANs' dynamical system can be improved by designing custom controllers from the control theory. 
With some modifications to the dynamics, several works~\cite{Kodaliconvergestability, xu2019understanding} can achieve asymptotic stability. However, these works have limitations. 
The stability they achieve is \emph{local}, which assumes that the network is initialized  sufficiently close to an equilibrium. Such an assumption cannot be guaranteed in practice. Furthermore, they only obtain \emph{asymptotic} stability, without a guarantee to the convergence rate.

In this work, we leverage potent tools of noise-based controllers from control theory to propose a \emph{Brownian motion controller} (BMC) to stabilize GANs' training. 
BMC stabilizes the training dynamics by adding noise, and the new noise-induced dynamics can be formulated as a system of stochastic differential equations (SDEs). 

We extensively analyze BMC for Dirac-GANs~\cite{mescheder2018training}, a popular class of GANs for analysis, to derive bounds on the rate of convergence. For Dirac-GANs, BMC are able to converge \emph{globally} with \emph{exponential} stability. Such theoretical result is much stronger than previous \emph{local}, \emph{asymptotic} ones~\cite{Heuse2017localnash,Kodaliconvergestability, xu2019understanding}. Our global converge result implies that the trajectory can converge with \emph{any} initialization rather than those near equilibria. Our globally exponential stability result further guarantees a fast convergence rate. 

Empirically, BMC improves the stability of the training of several popular GAN architectures, including classical DCGAN~\cite{radford2015dcgan} and the advanced StyleGANv2-ada~\citep{karras2020training}, which already has a carefully designed architecture and path-length regularization for stabilization. We evaluate BMC on a wide variety of datasets including CIFAR-10, LSUN-Bedroom 256x256, and FFHQ 1024x1024. Experiment results indicate that BMC can accelerate the training with a smaller range of oscillation, significantly reduced the training time, and achieved better image quality in terms of FID score.

\begin{figure}[tp!]
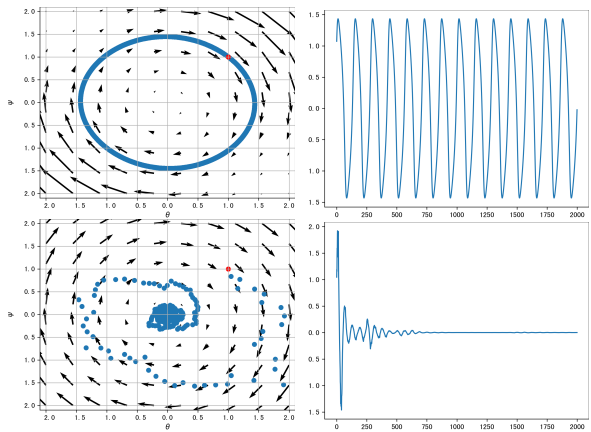
 
\begin{center}
\includegraphics[width=1.5in]{img/standard_gan-conv.pdf}
\includegraphics[width=1.5in]{img/standard_gan-grad.pdf}
\includegraphics[width=1.5in]{img/brownian_gan-conv.pdf}
\includegraphics[width=1.5in]{img/brownian_gan-grad.pdf}
\end{center}
\caption{The gradient map (left) and convergence behavior of generator parameter $\theta$ (right) of Dirac-WGAN (first row) and DiracWGAN-BMC (second row), where the Nash equilibria of both models are at $(0,0)^\top$.}
\label{dirac-figure}
\end{figure}

\section{Related Work}
In this section, we discuss previous attempts for stabilizing GANs and present some background of noise-based controllers in control theory. We compare the theoretical guarantee of stability of some representative works in Tab.~\ref{tab:my_label}.
\paragraph{Alternative Training Methods or Model Architecture}
To stabilize GANs' training process, a lot of work has been done on modifying its architecture. \citet{wang2021spatial} observe that during training, the discriminator converges faster and dominates the dynamics. They produce an attention map from the discriminator and use it to improve the spatial awareness of the generator. In this way, they claim to push GANs' solution closer to equilibrium. \citet{Heuse2017localnash} use a two-time scale update rule (TTUR) to push GANs' training process converging to a local Nash equilibrium. \citet{Karras2018progressive} train the generator and the discriminator progressively to stabilize the training process. Later on, StyleGANs models \citep{karras2019style,karras2020training,karras2020analyzing, karras2021alias} use style transfer to alternate the architecture for the generator and accomplish the state-of-the-art performance on high-resolution image synthesis. However, their methods do not have guarantees of global stability.

\paragraph{Regularization on Objective Functions}
Many works stabilizes GANs' training process with modified objective functions. \citet{Kodaliconvergestability} add gradient penalty to their objective function to avoid local equilibrium with their model called DRAGAN. This method has fewer mode collapses and can be applied to a lot of GANs frameworks. Other work, such as Generative Multi-Adversarial Network (GMAN) \citep{Durukar2017multi}, packing GANs (PacGAN) \citep{Lin2017Pac}, GANs with spectrum control \cite{jiang2019computation}, and energy-based GANs \citep{Zhao2016energy},  modifies the discriminator to achieve better stability. However, in their works, neither local nor global stability is promised and thus is orthogonal to our approach.  

\paragraph{Other Methods with Control Theory}
\citet{xu2019understanding} formulate GANs as a system of differential equations and add closed-loop control (CLC) on a few variations of the GANs to enforce stability. However, the design of their controller depends on the objective function of the GANs models and does not work for all variations of the GANs models. Additionally, their method is neither globally nor exponentially stable, which we find unable to stabilize later proposed StyleGANv2-ada \citep{karras2020training}. 

\paragraph{Dynamic Systems with Noise-based Controller}
In control theory, theoretical works have proved that white noise is able to stabilize linear \citep{arnold1983stabilization, scheutzow1993stabilization} and non-linear \citep{mao1994stochastic} dynamic systems. Later on, numerical experiments by \citet{toral2001analytical, he2003noise, lin2006using,sun2011generating}  observe a phenomenon called \textit{Noise-induced synchronization}, which indicates adding designed independent Gaussian white noise successfully stabilizes dynamic systems. Our BMC incorporates the idea of noise-induced synchronization, in which we analyze GANs' training process from control theory's perspective and design an invariant Brownian Motion Controller (BMC) to stabilize GANs' training process.

%% file: data/2-gan.tex
\section{Formulating GANs' Training as a Dynamical System}

In this section, we briefly review the formulation of GAN training as a stability problem of dynamical systems. 
Generative adversarial networks (GANs) fit a data distribution $p(x)$ with a generator $G$ and a discriminator $D$. The objective functions of GANs can be written as the following optimization problem:
\begin{equation}\label{ganobjective}
\resizebox{.98\hsize}{!}{$
\left\{
  \begin{array}{ll}
    \displaystyle\max_D L_D(D;G)= {\mathop{\mathbb{E}}}_{p(x)}[h_1(D(x))] + {\mathop{\mathbb{E}}}_{p_{G}(x)}[h_2(D(x))] \\
    \displaystyle\max_G L_G(G;D) = {\mathop{\mathbb{E}}}_{p_z(z)}[h_3(D(G(z)))], 
  \end{array}
\right.
$}
\end{equation}
where $h_1(\cdot)$, $h_3(\cdot)$ are increasing functions and $h_2(\cdot)$ is a decreasing function around zero. $p_{G}(x)$ is the distribution of the generator and $p_z$ is a Gaussian distribution.


GANs seek for a \emph{Nash equilibrium} $(D_*, G_*)$, where 
\begin{align*}
\forall D, L_D(D; G_*)\le L_D(D_*; G_*),\\
\forall G, L_G(G; D_*)\le L_G(G_*; D_*).
\end{align*}
However, finding the global Nash equilibrium is challenging. Gradient-based methods are usually adopted for Eq.~(\ref{ganobjective}). A convenient way to analyze GANs is thus to directly consider the dynamics defined by the optimization algorithm.

Following \citet{xu2019understanding}'s notation, the training dynamic (under  continuous time limit) for the generator and discriminator over the time domain $t$ can be written as: 
\begin{equation} \label{gan-differentialeq}
\begin{split}
\begin{aligned}
\resizebox{.99\hsize}{!}{$
    \left\{
  \begin{array}{ll}
    \displaystyle \frac{\textrm{d}D(x,t)}{\textrm{d}t}= p(x) \frac{\textrm{d}h_1(D(x,t))}{\textrm{d}D(x,t)}+p_G(x)\frac{\textrm{d}h_2(D(x,t))}{\textrm{d}D(x,t)}, \forall x & \hbox{} \\
    \displaystyle \frac{\textrm{d}G(z,t)}{\textrm{d}t}=p_z(z) \frac{\textrm{d}h_3(D(G(z,t)),t)}{\textrm{d}D(G(z,t),t)}\frac{\textrm{d}D(G(z,t),t)}{\textrm{d}G(z,t)}, \forall z\\
  \end{array}
\right.$
}
\end{aligned}
\end{split}
\end{equation}
where $D(x, t)$ and $G(z, t)$ are respectively the generator and the discriminator over a time domain $t$.
Define the state $X(t) = (D(x,t), G(z,t))^\top$ and the transition function
\begin{equation*}
    f(X(t)) =  
    \begin{pmatrix}
    p(x) \frac{\textrm{d}h_1(D(x,t))}{\textrm{d}D(x,t)}+p_G(x)\frac{\textrm{d}h_2(D(x,t))}{\textrm{d}D(x,t)} 
    \\
    p_z(z) \frac{\textrm{d}h_3(D(G(z,t),t))}{\textrm{d}D(G(z,t),t)}\frac{\textrm{d}D(G(z,t),t)}{\textrm{d}G(z,t)} 
    \end{pmatrix} ,
\end{equation*}
we can rewrite Eq.~(\ref{gan-differentialeq}) as 
\begin{equation} \label{norm_system}
    \textrm{d}X(t)=f(X(t))\textrm{d}t.
\end{equation}

Rather than Nash equilibria, we analyze whether the algorithm can reach an \emph{equilibrium of the dynamics} Eq.~(\ref{norm_system}), i.e., a point $X_e$ where $f(X_e)=0$, which we refer by ``equilibrium'' in the rest of the paper. Such an equilibrium cannot be improved by local gradient updates, and is closely related to the concept of \emph{local Nash equilibrium}~\cite{Heuse2017localnash}. 

If there exists an equilibrium, the question of interest is 
\begin{center}
\emph{Can the training algorithm find such an equilibrium?}
\end{center}
This closely relates to the concept of \emph{stability} in dynamical system theory, defined as follows:
\begin{definition}
 (asymptotic stability) Given a dynamic system of with equilibirum $X_e$, this system is said to be \textit{asymptotically stable}, if there exists a $\delta > 0$, such that whenever $\norm{X(0)- X_e} \leq \delta$, we have $\lim_{t\to\infty} \norm{X(t) - X_e} = 0 $.
\end{definition}
Intuitively, an equilibrium $X_e$ of a system is stable if the system can converge to  $X_e$ given any initial point close enough to it.
Here, the convergence radius $\delta$ is of practical interest. 
While several works~\cite{Heuse2017localnash,xu2019understanding,sonderby2016amortised} show that (possibly modified) GAN training dynamics are stable, the concept of stability only guarantees convergence in a small neighborhood. However, the practical implication of such theoretical results is questionable since we cannot guarantee to initialize the network in the vicinity of an equilibrium. Additionally, in practice, we are concerned about the training time required for GANs to converge. However, current theoretical 
results only guarantee convergence as time goes to infinity.

%% file: data/3-preliminary.tex
\section{Brownian Motion Controller for GANs}
Viewing GANs' training as a dynamic system, we can leverage advances in control theory to design novel controllers to stabilize the dynamics. The controllers define a new dynamical system with identical equilibria to the original system Eq.~(\ref{norm_system}), but with better stabilization properties.

Before presenting formal theoretical results, we notice that while existing work \citep{mescheder2018training,xu2019understanding} extracts GANs' training dynamic as a system of ordinary differential equations (ODEs), our new dynamics Eq.~(\ref{2.2}) is a system of stochastic differential equations (SDEs) due to the noise terms. 
Here, we present some preliminary concepts on SDEs and generalize the concept of stability to SDEs. 


\subsection{Preliminary}
\subsubsection{A System of SDEs}

In statistical physics and engineering, the motion of many systems can be  expressed as a SDE called \textit{Langevin Equation} \citep{kloeden1992stochastic} such that
\begin{equation} \label{br_ctrl}
    \tfrac{\textrm{d}X(t)}{\textrm{d}t} = f(X(t)) + g(X(t))B(t), 
\end{equation}
where $X(t)$ is a $m$-dimensional vector representing the state of the system and $B(t) = (B_1(t), \hdots, B_m(t))^T$ is a $m$-dimensional \emph{Brownian motion}, where each one-dimensional Brownian motion $\{B_i(t)\}_{t\ge 0}$ is a real-valued process  with the following properties \citep{mao2007stochastic}:
\begin{myitems}
    \item $B_i(0) = 0$ a.s.;
    \item for $0 \leq s < t< \infty$, the increment $B_i(t) - B_i(s)$ is normally distributed with mean 0 and variance $t-s$;
    \item The increments of the above property are independent to each other.
\end{myitems}

Given an initial state $X_0$, the training trajectory of GAN  can be viewed as a solution of an initial value problem.
\begin{definition} \citep{ito1951stochastic}
    Given a system described with the SDEs Eq.~(\ref{br_ctrl}) with the initial value $X(0) = X_0$, a \textit{solution $\{X(t)\}_{t\geq 0}$} to this system is a stochastic process  such that for $0\leq t \leq \infty$, we have
\begin{equation}
    X(t) = X_0 + \int_0^t f(X(s))ds + \int_0^t g(X(s))dB(s).
\end{equation}
\end{definition}

For any given system, we are concerned about whether this system exists a solution for a given initial point, and if so, whether the solution is unique \citep{mao1991note}. While for ODE, the existence and uniqueness of an initial value problem is guaranteed by the Picard–Lindel\"{o}f theorem, extra care needs to be taken for SDEs. We give the definitions below:
\begin{definition} \citep{veretennikov1981strong}
    A system of SDEs \textit{exists a global solution} if for any initial value $X_0$, its solution $\{X(t)\}_{t\geq 0}$ does not diverge and approach vertical asymptotic (line $x = a$ for some a) on $t \in [0,\infty)$.
\end{definition}

\begin{definition} \citep{mao2006stochastic}
    A system of SDEs has \textit{a unique solution} if given \emph{any} initial value $X(0) = X_0$, there almost surly exists one and only one solution $\{X(t)\}_{t\geq 0}$ starting with this initial value.
\end{definition}

\subsubsection{Stability Analysis}
Modern stability theory of dynamical systems is developed under Lyapunov's second method \citep{goldhirsch1987stability, shevitz1994lyapunov, sastry1999lyapunov}. Throughout our work, we evaluate the stability of GANs' training process under the framework of Lyapunov stability analysis. First, we generalize the definition of equilibria to SDEs:
\begin{definition}
    $X_e$ is said to be an \textit{equilibrium }  of a system of SDE, if there exists a $t_0$ such that $X(t_0) = X_e$ and for all $t \geq t_0$ satisfies 
        $\mathbb{P}\big(f(X(t)) + g(X(t))B(t) = 0 \big) = 1.$

\end{definition}
As that for ODEs,  equilibrium implies for a state which cannot be improved with the stochastic updates Eq.~(\ref{br_ctrl}) and relates to a local Nash equilibrium. Then, whether the solution can converge to an equilibrium relates to the following definitions of stability:
\begin{definition} \label{stability}
 Given a system of SDEs with an initial condition $X(0) = X_0$ and equilibrium $X_e$, if we assume $\{X_t\}_{t \geq 0}$ is its unique solution. This system is said to be
    \begin{myitems}
    \item \textbf{Lyapunov stable}, if given $\epsilon > 0$, there exists a $\delta > 0$ such that whenever $\norm{X(0)- X_e} \leq \delta$, we have $\mathbb{P}\big(\norm{X(t) - X_e)} < \epsilon \big) =1$ for $0 \leq t \leq \infty$.
    \item \textbf{asymptotically stable}, if this system is Lyapunov stable and there exists a $\delta>0$, such that whenever $\norm{X(0)- X_e)} \leq \delta$, we have $\mathbb{P}\big(\lim_{t\to\infty} \norm{X(t) - X_e} = 0 \big) = 1$.
    \item \textbf{exponentially stable}, if it is asymptotically stable and there exists $\alpha > 0, \beta > 0, \delta > 0$ such that whenever $\norm{X(0)- X_e} \leq \delta$, for $0 \leq t \leq \infty$, we have $\mathbb{P}\big(\norm{X(t) - X_e} \leq \alpha \norm{X(0) - X_e}e^{-\beta t}\big) =1 $.
    \item \textbf{unstable}, if neither of the three conditions above is satisfied.
\end{myitems}
\end{definition}

In what we follow, we say a dynamic system is \textit{\textbf{globally}  Lyaponouv/ asymptotically/ exponentially stable}, if for \emph{any} initial value, this system has a unique solution converging to equilibrium and exhibits Lyaponouv/ asymptotically/ exponential stability. Under this circumstance, the $\delta$ in definition \ref{stability} can be arbitrary, leading to stronger stability than what we defined in definition \ref{stability}.




\subsubsection{Noise-based Control on ODEs}
Given a system of ODEs $\frac{\textrm{d}X(t)}{\textrm{d}t} = f(X(t))$ with an equilibruim $X_e$
which exihibts unstable nature, we say the controller $g(X(t))B(t)$ globally Lyapunov / asymptotically / exponentially \textit{stabilizes} the system $\frac{\textrm{d}X(t)}{\textrm{d}t} = f(x(t))$
if and only if $X_e$ is still an equilibruim and the system of SDEs $\frac{\textrm{d}X(t)}{\textrm{d}t} = f(X(t)) + g(X(t))B(t)$ is globally Lyapunov / asymptotically / exponentially stable.

\begin{theorem} \citep{mao1994stochastic}
 A system of ODEs $\frac{\textrm{d}X(t)}{\textrm{d}t} = f(X(t))$ with $X(0)=X_0$ can be almost surely exponentially stabilized by some noise term $g(X(t))B(t)$ if their exists $K > 0$, such that $f(X(t))$ satisfies \\
    \begin{equation}
        \abs{f(X(t))} \leq K\norm{X}, \forall X \in \mathbb{R}^m, t \geq 0
    \end{equation}
\end{theorem} 

\subsection{Brownian Motion Controller}
In control theory, noise-based controllers are useful tools for stabilizing dynamical systems and pushing the solution towards optimal value over time domain $t$ \citep{br2002}. 

In our work, we propose \emph{Brownian motion controller} (BMC), a higher order noise-based controller as a universal control function invariant to objective functions of GANs. 
Eq.~(\ref{norm_system}) is a second-order dynamical system. We follow the framework of controllers for high order system proposed by \citep{wu2009suppression} to propose our BMC as below:    
\begin{equation}\label{control2}
    u(t)=\varrho_{1}X(t)\dot{B}_{1}(t)+\varrho_{2}|X(t)|^{\beta}X(t)\dot{B}_{2}(t),
\end{equation}
where $B_{1}(t)$ and $B_{2}(t)$ are independent one-dimensional Brownian motions, $\varrho_{1}$ and $\varrho_{2}$ are non-negative constants.
Incorporating BMC Eq.~(\ref{control2}), the controlled system becomes
\begin{equation}\label{2.2}
    \textrm{d}X(t)=f(X(t))\textrm{d}t+ u(t).
\end{equation}
Intuitively, BMC adds multiplicative noise, i.e., where the strength of stochasticity depends on the current state of the system, to the original system. We will show in Sec.~\ref{section:dirac}-\ref{section:normal-gan} (theoretically) and Sec.~\ref{section:evaluation} (empirically) that training GANs with BMC can have better stabilization properties than previous GAN training dynamics and controllers~\cite{Kodaliconvergestability,xu2019understanding}. Importantly, on Dirac-GAN, a popular class of GANs for analysis, BMC is \emph{globally exponentially stable}. This means the training process can converge \emph{exponentially fast} to the equilibrium starting from \emph{any} initialization point. This theoretical result is much stronger than previous ones, which are only \emph{locally asymptotically stable}.

%% file: data/4-dirac-gan.tex
\section{Stability Analysis for Dirac-GANs}\label{section:dirac}
In this section, we focus on Dirac-GANs, a simplified GANs' setting with linear generator and discriminator networks proposed by \citet{mescheder2018training} to illustrate the unstable nature of GANs. We prove that Dirac-GAN with BMC is globally exponentially stable, and we derive bounds on its converge rate. 


\subsection{Dynamic System of Dirac-GANs}
In Dirac-GANs' settings, the generator is a point mass following $p_G(x) = \delta(x- \theta)$ and the discriminator is a linear $D_{\phi}(x) = \phi x$. The true data distribution is given by $p_D(x) = \delta(x-c)$ with a constant c. Here, $\delta(\cdot)$ is a Dirac distribution with $\delta(x)=0$ for $x\ne 0$ and $\int_{-\infty}^{\infty} \delta(x)=1$.


The objective functions of Dirac-GANs can be written as:
\begin{equation} \label{eqn:dirac-gan}
\left\{
  \begin{array}{ll}
    \displaystyle \max_\phi L_{D_\phi}(\phi; \theta)= h_1(D_\phi(c)) + h_2(D_\phi(\theta)) \\
    \displaystyle \max_\theta L_{G_\theta}(\theta;\phi) = h_3(D_\phi(\theta)),\\
  \end{array}
\right.\
\end{equation}
where $h_{1}(\cdot)$ and $h_{3}(\cdot)$ are increasing functions and $h_{2}(\cdot)$ is a decreasing function around zero \citep{xu2019understanding}. 





Taking the reparameterization $\tilde{\theta}(t)=\theta(t)-c$, we have the following dynamic system with a unique equilibrium $(0,0)^{T}$:
\begin{equation}\label{14}
\resizebox{.99\hsize}{!}{$
\left\{
  \begin{array}{ll}
    \displaystyle \tfrac{\textrm{d}\phi(t)}{\textrm{d}t}= h^{\prime}_1(\phi(t)c)c+h^{\prime}_2(\phi(t)(\tilde{\theta}(t)+c))(\tilde{\theta}(t)+c) & \hbox{} \\
    \displaystyle \tfrac{\textrm{d}\tilde{\theta}(t)}{\textrm{d}t}= h^{\prime}_3(\phi(t)(\tilde{\theta}(t)+c))\phi(t).\\
  \end{array}
\right.$}
\end{equation}

Depending on the specific choice of $h_1$, $h_2$, and $h_3$, the training dynamics of many GANs can be represented by Eq.~(\ref{14}) with the minimalistic generator and discriminator architecture. These GANs are referred as Dirac-GAN, Dirac-WGAN, and so on. Unfortunately, \citet{mescheder2018training} find that many Dirac-GANs are unstable. Our experiments also confirm the unstable nature of Eq.~(\ref{14}). As in Fig.~\ref{dirac-figure}, Dirac-WGAN is unstable and circles around the origin.

To stabilize Eq.~(\ref{14}), we apply our designed BMC and transform Dirac-GANs' training dynamics to Eq.~(\ref{2.2}), 
where $X(t) = (\phi(t), \theta(t))^\top$, $f(X(t)) = (h^{\prime}_1(\phi(t)c)c+h^{\prime}_2(\phi(t)(\tilde{\theta}(t)+c))(\tilde{\theta}(t)+c), h^{\prime}_3(\phi(t)(\tilde{\theta}(t)+c))\phi(t))^\top$, and $u(t)$ is defined as in Eq.~(\ref{control2}).

\subsection{Dirac-GANs are  Exponentially Stable with BMC}
In this section, we derive the existence of unique global solution and stability of system (\ref{2.2}). For the stability analysis, we impose the following assumption on the smoothness of functions $h_{1}$, $h_{2}$, $h_{3}$ in system (\ref{2.2}).

\begin{assumption}\label{ass2}
There exist positive constants $\alpha_{1}$, $\alpha_{2}$, $\alpha_{3}$ such that for any $x,y\in {\mathbb{R}}^n $,
\begin{equation*} 
\resizebox{.99\hsize}{!}{$
    |h^{\prime}_1(x)-h^{\prime}_1(y)|\leq \alpha_{1}||x-y||,~~|h^{\prime}_2(x)-h^{\prime}_2(y)|\leq \alpha_{2}||x-y||,$
    }
\end{equation*}
\begin{equation*}
    |h^{\prime}_3(x)-h^{\prime}_3(y)|\leq \alpha_{3}||x-y||.
\end{equation*}
\end{assumption}

In what follows, we first prove that the BMC from equation (\ref{control2}) yields a unique global solution in Theorem \ref{thm1}. Then, in Theorem \ref{thm2}, we show that this unique global solution exponentially converges to the equilibrium point $a.s.$ with bounds on the hyper-parameters $\varrho_{1}, \varrho_{2}$ and $\beta$ which in turn affect the rate of convergence. Combining Theorem \ref{thm1} and Theorem \ref{thm2}, we claim that with our BMC, Dirac-GANs representing by system (\ref{2.2}) is globally exponentially stable.

\begin{theorem} \label{thm1}
(Proof in Appendix A) Under Assumption \ref{ass2}, for any initial value $X(0)=\xi\in  \mathbb{R}^{2}$, if $\varrho_{2}\neq 0$ and $\beta>1$, then there a.s. exists a unique global solution $X(t)$ to system (\ref{2.2}) on $t\in[0, \infty)$.
\end{theorem}


\begin{theorem} \label{thm2}
(Proof in Appendix B) Let Assumption \ref{ass2} hold. Assume that $\varrho_{2}\neq 0$ and $\beta>1$. If
$
\frac{\varrho_{1}^{2}}{2}-\varphi>0,
$
where $\varphi$ takes the value of 
\begin{equation}\label{fai}
\max_{x\geq 0}\Bigg\{-\frac{\varrho_{2}^{2}}{2}x^{2\beta}+(\alpha_{2}^{2}+\frac{1}{2}\alpha_{3}^{2})x^{2}+[(1+\frac{1}{2}\alpha_{1}^{2})c^{2}
+2c+\frac{1}{2}]\Bigg\},
\end{equation}
then for any $X(0)=\xi$ with sufficiently small constant $\epsilon\in(0, \varrho_{1}^{2}/2-\varphi)$, the global solution $X(t)$ of system (\ref{2.2}) has the property that
\begin{equation*}
\begin{split}
\begin{aligned}
\limsup_{t\rightarrow\infty}\frac{\log|X(t)|}{t}\leq-\bigg(\frac{\varrho_{1}^{2}}{2}-\varphi\bigg)+\epsilon,~~~~a.s.
\end{aligned}
\end{split}
\end{equation*}
that is, the solution of system (\ref{2.2}) is a.s. exponentially stable.
\end{theorem}

Here since $\frac{\varrho_{1}^{2}}{2}-\varphi>0$ and $\epsilon$ is a sufficiently small constant, then when Eq. (\ref{fai}) is satisfied, we have 
$
\limsup_{t\rightarrow\infty}\frac{\log|X(t)|}{t}\leq -\lambda,~~~~a.s.
$
for some positive constant $\lambda$. Rearranging we get 
$
\limsup_{t\rightarrow\infty}|X(t)|\leq e^{-\lambda t},
$
which implies 
$
\limsup_{t\rightarrow\infty}X(t) = (0,0)^\top
$
as required. Notice that the rate of convergence depends only on constant $\lambda$, which in turn depends on $\varrho_1$ and $\varphi$. Thus the convergence rate is decided by the choice of hyper-parameters $\varrho_1, \varrho_2$, and $\beta$. In practice, we can tune these three variables as desired, as long as they satisfy the constraint from Eq.~(\ref{fai}).

Notice that our BMC work for any $h_1, h_2$ and $h_3$ as long as they satisfy the smoothness condition under assumption \ref{ass2}. In other words, we have proven that with BMC,  Dirac-GANs are \textit{globally exponentially stable} regardless of $h_1$, $h_2$ and $h_3$, and we have given theoretical bounds on the rate of convergence. In Fig.~\ref{dirac-figure}, we present visual proof that the Dirac-GAN with BMC is stable and converges to the optimal equilibrium as required. 


%% file: data/5-normal-gan.tex
\section{Stabilize General GANs with BMC}\label{section:normal-gan}
In this section, we present BMC for general GANs beyond Dirac-GAN. 
Under GANs' setting, the generator and discriminator learn to compete with each other, and ideally, they would converge to their global optimal solution simultaneously. However, in reality, the generator and discriminator typically learn at different paces \citep{wang2021spatial}. For example, in DCGAN~\cite{radford2015dcgan} the discriminator learns faster, while in StyleGANv2~\cite{karras2020training} the generator learns faster. This inconsistency would result in the faster network being stuck in its local equilibrium while the slower one stops learning useful information and lead to mode-collapses or unstable behaviors.

BMC considers both the generator and discriminator as a \textit{coupled dynamic system} (a system of two differential equations with two dependent variables $D$ and $G$, and one independent variable $t$) \citep{peng1999fully} and the goal of our BMC is to make sure the generator and discriminator learn at the same pace so that they are able to converge to optimal equilibrium at the same time.



For simplicity, we take $\beta = 2$ (the order of noise often take an even number) for general GANs, then BMC in this setting becomes
\begin{equation} \label{control_normal_gan}
    u(t)=\varrho_{1}X(t)\dot{B}_{1}(t)+\varrho_{2}|X(t)|^{2}X(t)\dot{B}_{2}(t)
\end{equation}

Our BMC can be reflected on objective functions as a regularization term on both generator and discriminators, with the derivative being Eq.~(\ref{control_normal_gan}). We thus take integration of Eq.~(\ref{control_normal_gan}) and modify the objective functions in (\ref{ganobjective}) to:

\begin{equation} \label{brgan_obj}
\resizebox{.99\hsize}{!}{$
\begin{aligned}
    \left\{
  \begin{array}{ll}
    \max_D L_D{'}(D;G) = L_D(D;G) + \frac{1}{2}\varrho_{1} D^2(x) \dot{B}_{1}(t)\\
    + [\frac{1}{4}\varrho_{2}D^4(x) + \frac{1}{2}\varrho_{2}D^2(G(z))D^2(x)]\dot{B}_{2}(t). \\
    \max_G L_G{G}(G;D) = L_G(G;D) + \frac{1}{2}\varrho_{1} D^2(G(z)) \dot{B}_{1}(t)\\
    + [\frac{1}{4}\varrho_{2}D^4(G(z)) + \frac{1}{2}\varrho_{2}D^2(G(z))D^2(x)]\dot{B}_{2}(t).\\
  \end{array}
\right.
\end{aligned} $}
\end{equation}

Although in general GANs' setting, we are not aware of the equilibrium point of the generator, according to Theorem~\ref{thm2}, this equilibrium would affect the constraints between $\varrho{1}$ and $\varrho_{2}$. As a result, the stability is only guaranteed for certain pairs of $\varrho{1}$ and $\varrho_{2}$. We implement our designed objective functions in Sec.~\ref{section:evaluation} and analyze results for various pairs of $\varrho{1}$ and $\varrho_{2}$. Our numerical experiments show that BMC successfully stabilizes GANs models and are able to generate images with promising quality.

%% file: data/6-experiments.tex
\section{Evaluation} 
\label{section:evaluation}
\begin{figure*}[htbp!] 
    \centering
    \includegraphics[scale=0.45]{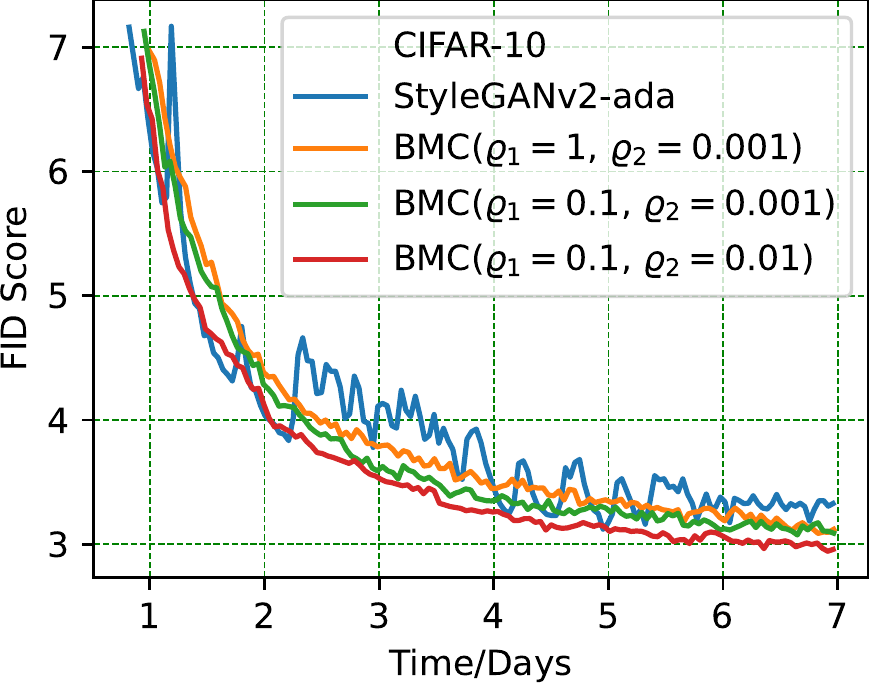}
    \includegraphics[scale=0.45]{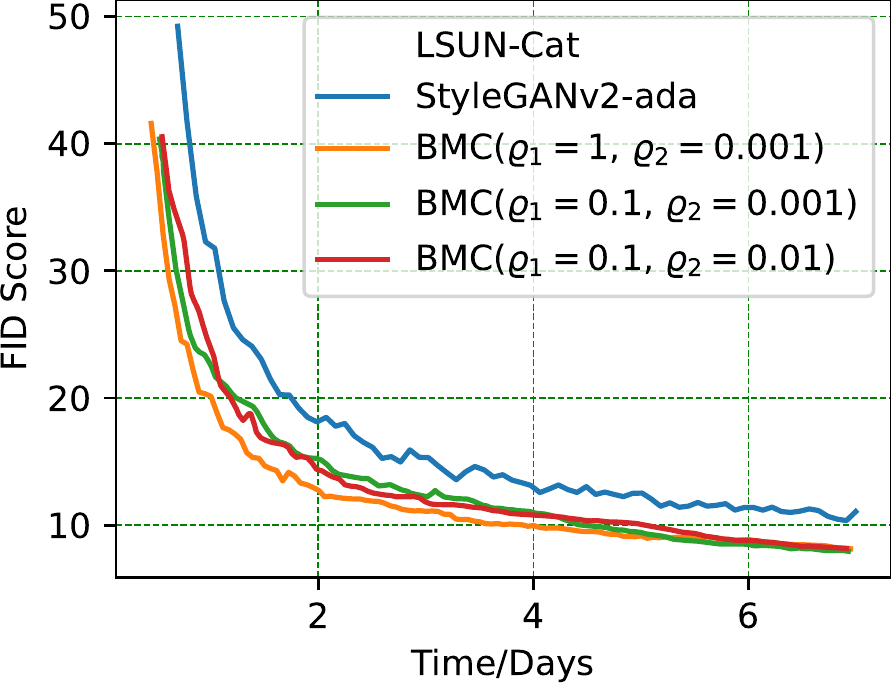}
    \includegraphics[scale=0.45]{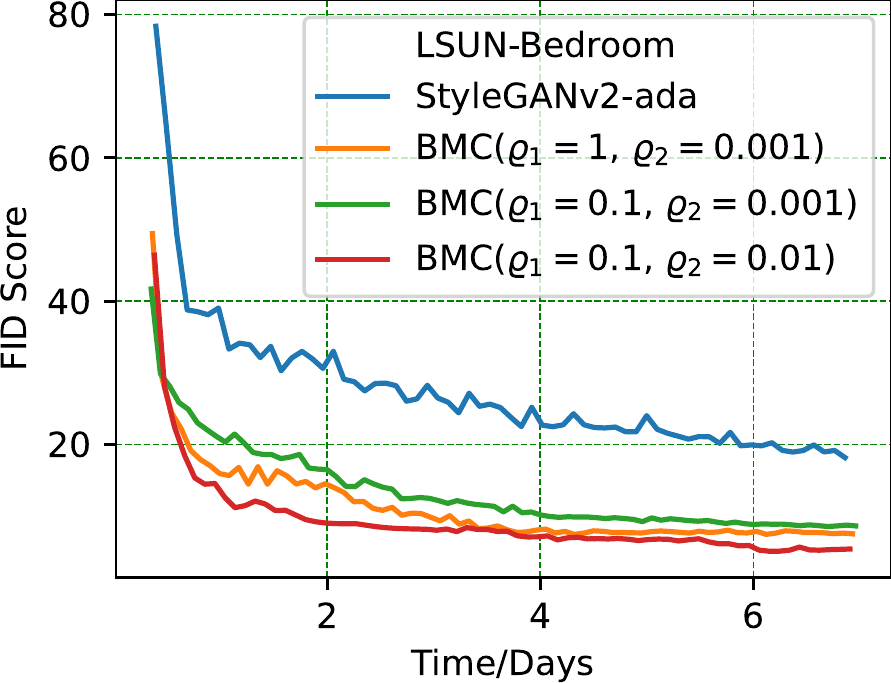}
    \includegraphics[scale=0.45]{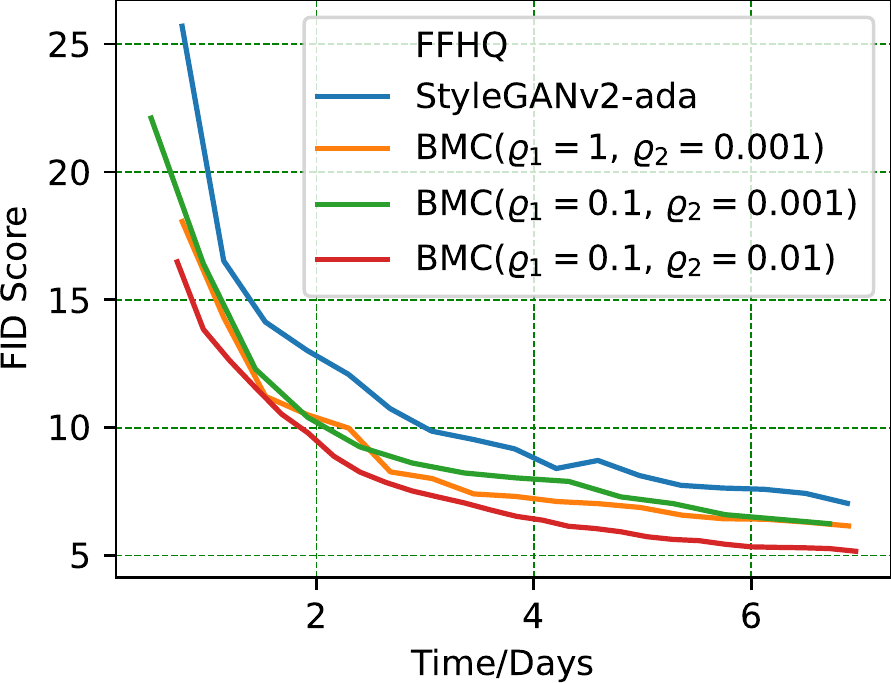}
    \caption{The FID score curves with respect to time. The blue curve represents StyleGANv2-ada baseline using the same parameter settings as their original paper \citep{karras2020training} , while the green, orange and red curves add BMC and are evaluated under the same setting.}
    \label{style_fid}
\end{figure*}

In this section, we show the effectiveness of BMC by providing both quantitative and qualitative results. We first evaluate the performance of BMC for Dirac-GANs (``Dirac-GAN-BMC'')  in Sec.~\ref{eval_dirc} and present Fig.~\ref{dirac-figure} to illustrate that our BMC successfully stabilizes Dirac-GANs. Additionally, we compare the number of iterations needed for Dirac-GAN-BMC to converge under various parameters of BMC. Then in Sec.~\ref{eval_normal}, we compare our GAN-BMC with the StyleGANv2-ada baseline \citep{karras2020training} on multiple datasets with various parameter settings of BMC. Experiments show that our BMC effectively stabilizes the training dynamics by reducing the range of oscillation, speeding up the convergence rate, and improving the FID score.

\subsection{Convergence of Dirac-GANs with BMC} \label{eval_dirc}
In Dirac-GAN's setting, we know the optimal equilibrium is $(0,0)^\top$, so we can measure the convergence speed and draw the gradient map for comparison, which are presented in Fig.~\ref{dirac-figure}. These results show that our Dirac-GAN-BMC has better convergence patterns and speed than Dirac-GANs. Without adding BMC to the training objective, Dirac-GANs cannot reach equilibrium. The parameters of the generator and the discriminator oscillate in a circle, as shown in Fig.~\ref{dirac-figure}. In comparison, these parameters of the Dirac-GAN-BMC only oscillate in the first 500 iterations and soon converge in 800 iterations.

We also study different combinations of the hyperparameters $\varrho_{1}$ and $\varrho_{2}$ under $\beta=1$ and $\beta=2$. As shown in Tab.~\ref{tabb1}, Dirac-GANs-BMC converge better when we set $\varrho_{1}=0.1$ and $\varrho_{2}=0.01$. Generally, a larger $\varrho$ will lead to a faster convergence rate, but when $\varrho$ is large enough, the effect of increasing $\varrho$ will become saturated. On the other hand, when $\varrho$ is too small, Dirac-GAN-BMC will take more than 100000 iterations to converge.

\begin{table}[!htpb]
\centering
\caption{Converge iterations required for $\beta=1$ and $\beta=2$  on DiracGANs-BMC. Symbol N below indicates not converging after 100k iterations.}
\label{tabb1}
\begin{tabular}{llll}
\toprule[1.2pt]
$\beta=1/2$           & $\varrho_{2}=0.0001$ & $\varrho_{2}=0.001$  & $\varrho_{2}=0.01$ \\
\midrule
$\varrho_{1}=0.1$   & 0.6k / 1.5k                 & 0.4k / 0.75k                  & 0.4k / 0.7k                \\
$\varrho_{1}=0.01$  & 25k / N                & 15k / 9k               & 10k  / 8.5k            \\
$\varrho_{1}=0.001$ & N / N & N / N & 40k / N              \\ 
\bottomrule[1.2pt]
\end{tabular}
\end{table}

 

\subsection{GANs-BMC: Converge better and lower FID} \label{eval_normal}
\textbf{Dataset:} We evaluate our proposed GANs-BMC on well-established CIFAR10 \citep{krizhevsky2009learning}, LSUN-Bedroom with resolution 256x256 \citep{yu2015lsun}, LSUN-Cat with resolution 256x256 \citep{yu2015lsun}, and FFHQ with resolution 1024x1024 \citep{karras2019style}. 


\textbf{Implementation details:}
We compare the FID score \citep{heusel2017gans} of StyleGANv2-ada \citep{karras2020training} and its stabilized version with our BMC (``StyleGANv2-ada-BMC''). We reproduce the identical configuration settings as reported in the StyleGANv2-ada paper within the period of 7 days on 4 cards of NVIDIA GeForce GTX TITAN X. The detailed experimental setups can be found in Appendix C. We find that adding BMC results in better performance without changing any hyperparameters of the network, as shown in Fig.~\ref{style_fid} . We conduct multiple trials under different coefficients of BMC and report the FID score with its range of oscillation in Tab.~\ref{para}. Additionally, we calculate the shifting difference of the generator throughout the training process, shown in Fig.~\ref{fig:difference}.


\subsection{Quantitative Results}
As in Fig. \ref{style_fid} and Tab. \ref{para}, StyleGAN2-ada has a large range of oscillation on FID scores. Especially on the low-resolution dataset CIFAR-10, we observe that the FID curve oscillates even after training for a long time. On the other hand, StyleGANv2-ada-BMC is able to reduce this range of oscillation by a multiple of 10 in CIFAR-10 32*32, LSUN-Bedrrom 256*256, LSUN-Cat 256*256 and a multiple of 4 on FFHQ 1024*1024. As a result, BMC largely reduces the range of oscillation, providing better convergence behavior. 

\begin{table}[htpb!]  
\caption{FID scores for each dataset after training for 7 days. ($a \pm b$ in this table should be interpreted as $a$ is the FID score after 7 days, $b$ is the range of oscillation on the 24 hours span of the seventh day). }
\begin{adjustbox}{width={80mm},keepaspectratio}%
\begin{tabular}{|c|c|clc|}
\hline
\multicolumn{2}{|c|}{} &\multicolumn{1}{c|}{$\varrho_1 = 1$} & \multicolumn{1}{l|}{$\varrho_1 = 0.1$} &  $\varrho_1 = 0.01$ \\ \hline

\multirow{4}{*}{\begin{tabular}[c]{@{}c@{}}Cifar10\\ (32x32)\end{tabular}}  & $\varrho_{2} = 0.01$ & $ 3.12 \pm 0.03$& $\mathbf{2.94 \pm 0.01}$ & $3.56 \pm 0.06$\\ 
  &  $\varrho_{2} = 0.001$ &  $3.05 \pm 0.02$   &  $2.95\pm0.02$  & $3.31 \pm 0.04$  \\
 &   $\varrho_{2} = 0.0001$  &  $2.98 \pm 0.02$ & $2.96 \pm 0.02$  & $3.15 \pm 0.05$ \\ \cline{2-5} 
  & Baseline     & \multicolumn{3}{c|}{$3.32 \pm 0.16$}             \\ \hline

\multirow{4}{*}{\begin{tabular}[c]{@{}c@{}}LSUN\\Bedroom\\ (256x256)\end{tabular}} &$\varrho_{2} = 0.01$   & $\mathbf{5.34 \pm 0.09}$  & $5.40 \pm 0.08$  & $7.06\pm 0.09$ \\
 &  $\varrho_{2} = 0.001$  &    $7.52 \pm 0.07$    &   $8.62 \pm 0.12$             &  $8.82 \pm 0.15$\\ 
   & $\varrho_{2} = 0.0001$  & $7.48 \pm 0.06$  &  $9.01 \pm 0.14$ & $9.25\pm 0.17$ \\ \cline{2-5} 
  & Baseline     & \multicolumn{3}{c|}{$18.16 \pm 1.00$}             \\ \hline
               
\multirow{4}{*}{\begin{tabular}[c]{@{}c@{}}LSUN\\Cat\\ (256x256)\end{tabular}}     &  $\varrho_{2} = 0.01$   &    $8.42 \pm 0.06$       &           $8.07 \pm 0.04$         &  $ 9.09 \pm 0.16$\\ 
  &   $\varrho_{2} = 0.001$  & $8.03 \pm 0.04$      &  $\mathbf{8.01 \pm 0.03}$          &   $9.23 \pm 0.18$\\ 
  & $\varrho_{2} = 0.0001$  &     $9.56 \pm 0.23$   &    $9.30 \pm 0.21$        & $9.48 \pm 0.19$ \\ \cline{2-5} 
 & Baseline     & \multicolumn{3}{c|}{$10.57 \pm 0.31$}             \\ \hline
 
\multirow{4}{*}{\begin{tabular}[c]{@{}c@{}}FFHQ\\ (1024x1024)\end{tabular}} & $\varrho_{2} = 0.01$ &     $5.74 \pm 0.29$       &    $\mathbf{5.17 \pm 0.10}$            & $7.96\pm 0.14$ \\
&$\varrho_{2} = 0.001$  & $6.16 \pm 0.14$ &    $6.24 \pm 0.17$    & $7.32 \pm 0.18$  \\ 
&$\varrho_{2} = 0.0001$  &  $6.02 \pm 0.09$  & $6.03 \pm 0.11$        &  $8.01 \pm 0.23$\\ \cline{2-5}
& Baseline     & \multicolumn{3}{c|}{$7.75 \pm 0.42$}                \\ \hline
\end{tabular}
\end{adjustbox}
\label{para}
\end{table}

While StyleGANv2-ada's takes a long training time, especially on the dataset with high resolution, our BMC speeds up the training time for producing images with the identical FID scores by more than 2 days as observed in Fig.~\ref{style_fid}. 

\begin{figure}[htpb!]
    \centering
    \includegraphics[scale=0.5]{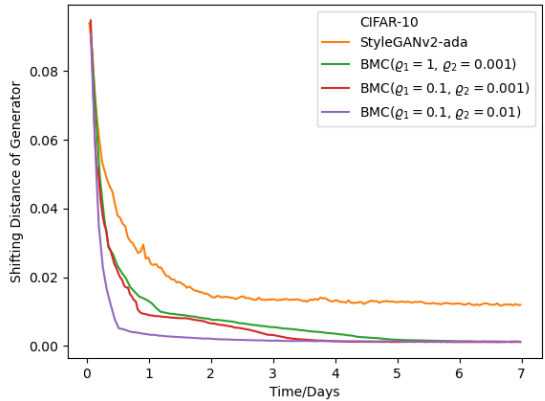}
    \caption{The shifting difference against time on CIFAR-10 baseline and its BMC trails. Clearly the shifting difference with BMC approaches 0.}
    \label{fig:difference}
\end{figure}

Additionally, we measure the training stability by defining the shifting difference of the generator as 
\begin{equation*}
    D(G_1, G_2) = {\mathop{\mathbb{E}}}_{p_z(z)} \norm{G_1(z) - G_2(z)}.
\end{equation*}
We calculate the shifting difference of the generator in between iterations and, as in Fig.~\ref{fig:difference}, while StyleGANv2-ada has a non-vanishing shifting difference at around 0.02, 
the shifting difference of our BMC approaches 0, indicating our BMC successfully stabilizes the generator.

\begin{figure}[h!]
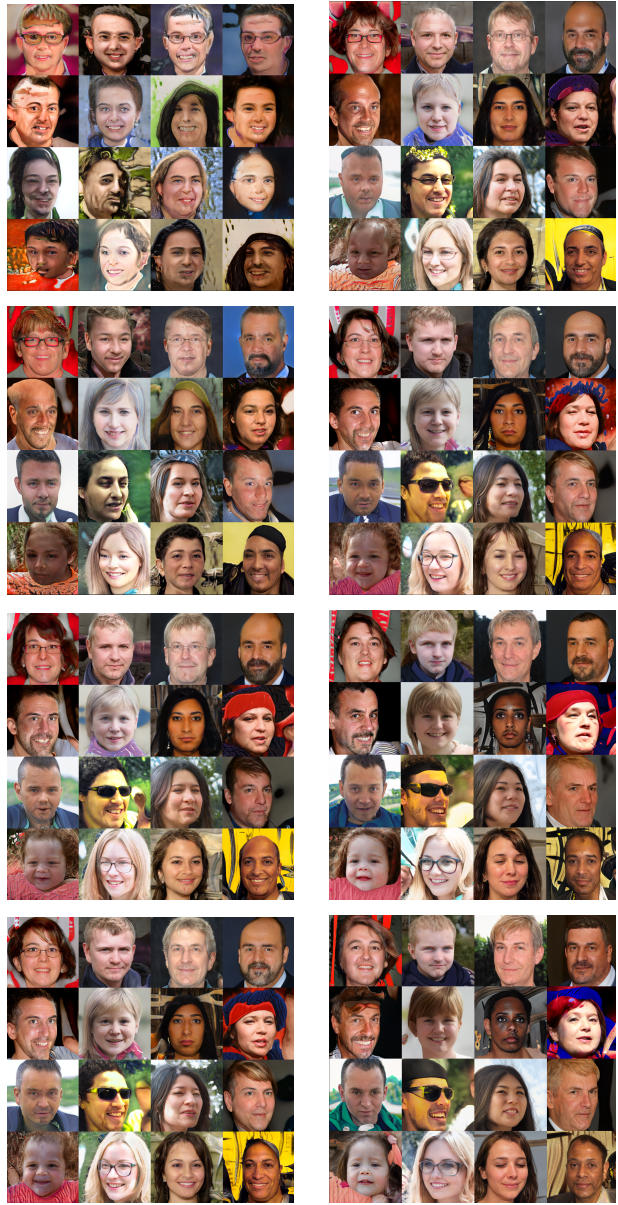
 
    \centering
    \begin{subfigure}{}
        \includegraphics[width=38mm]{img/ffhq-1.pdf}
    \end{subfigure}
    \hfill
    \begin{subfigure}{}
         \includegraphics[width=38mm]{img/ffhq-br-1.pdf}
    \end{subfigure}   

    \begin{subfigure}{}
        \includegraphics[width=38mm]{img/ffhq-2.pdf}
    \end{subfigure}
    \hfill
    \begin{subfigure}{}
        \includegraphics[width=38mm]{img/ffhq-br-2.pdf}
    \end{subfigure}

    \begin{subfigure}{}
        \includegraphics[width=38mm]{img/ffhq-3.pdf}
    \end{subfigure}
    \hfill
    \begin{subfigure}{}
        \includegraphics[width=38mm]{img/ffhq-br-3.pdf}
    \end{subfigure}

    \begin{subfigure}{}
        \includegraphics[width=38mm]{img/ffhq-4.pdf}
    \end{subfigure}
    \hfill
    \begin{subfigure}{}
        \includegraphics[width=38mm]{img/ffhq-br-4.pdf}
    \end{subfigure}

    \caption{Each row represents images generated from the FFHQ 1024*1024 dataset within the first 20 hours (5, 10, 15, and 20 hours). The first column of images is generated by StyleGANv2-ada, and the second column of images is generated after adding BMC.}
    \label{ffhq-img}
\end{figure} 

\subsection{Qualitative results}
We provide qualitative results on FFHQ 1024x1024 in Fig.~\ref{ffhq-img}. Noticing that our GANs-BMC is able to produce images with higher quality at a faster speed. 

